\documentclass[journal]{journal}

\usepackage{soul}
\usepackage{url}
\usepackage[hidelinks]{hyperref}
\usepackage[utf8]{inputenc}
\usepackage[small]{caption}
\usepackage{graphicx}
\usepackage{amsmath}
\usepackage{booktabs}
\usepackage{tikz}
\usetikzlibrary{positioning, fit, calc, matrix}
\usepackage{verbatim}
\newcommand{\name}{GRCNN}
\newcommand{\lcnote}[1]{\textcolor{blue}{#1}}
\newcommand{\ignore}[1]{}

\begin{document}

\title{\name{}:  Graph Recognition Convolutional Neural Network for Synthesizing Programs from Flow Charts}

\author{
Lin Cheng,
Zijiang Yang%
\thanks{L. Cheng and Z. Yang are with the Department of Computer Science, Western Michigan University, Michigan, USA (emails: lin.cheng@wmich.edu and zijiang.yang@wmich.edu).}
}

\maketitle
\thispagestyle{empty}

\begin{abstract}
  Program synthesis is the task to automatically generate programs
  based on user specification.
  In this paper, we present a framework that synthesizes programs from flow charts that serve 
  as accurate and intuitive specifications. In order doing so, we propose a deep neural network called \name{} that recognizes
  graph structure from its image. 
  \name{} is trained end-to-end, which can predict edge and node information
  of the flow chart simultaneously.
  Experiments show that the accuracy rate to synthesize a program is 66.4\%, and the accuracy rates to recognize edge and nodes are 94.1\% and  67.9\%, respectively. On average, it takes about 60 milliseconds to synthesize a program. 
\end{abstract}

\begin{IEEEkeywords}
program synthesis, flow chart, specification, graph recognition, CNN.
\end{IEEEkeywords}

\IEEEpeerreviewmaketitle

\section{Introduction}

\IEEEPARstart{P}{rogram} synthesis enables people to program computers without training in coding. It has been used  used in many domains such as data wrangling, graphs, and code repair \cite{gulwani2010dimensions}.
A good example is FlashFill \cite{gulwani2011automating}, which allows spreadsheet
users to provide a few examples and generates a program that conforms to the examples.

To synthesize a program,  specification must be provided.
Specification in formal language can accurately represent the user
intent and are used in deductive program synthesis \cite{manna1980deductive}.
However, very few have the
knowledge of formal language, so it cannot benefit most end users.
Under-specification is used in programming by example \cite{gulwani2011automating,feng2017component}, and
programming by demonstration \cite{lau1998programming}.
Under-specification does not require language knowledge and
is accessible by most end users.
However, because there may be more than one program that
satisfies the specification, how to choose the correct program
that captures the user intent is still an open problem.

\ignore{
As the progress of deep learning, researchers use deep learning techniques
to synthesize programs. 
Deepcoder, \cite{}, use
However, none of the deep learning techniques
synthesize programs with accurate specifications.
}

\begin{figure}
\centering
\input{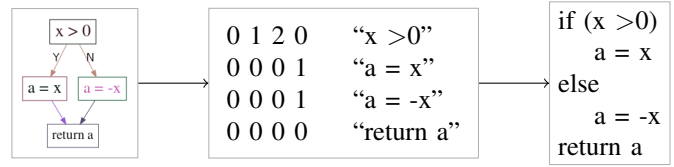}
\caption{Overview of \name{}. The input is the flow chart of \texttt{abs} function. The middle is the adjacent matrix and text of each node generated by  \name{}. The output is the synthesized source code.}
\label{fig:overview}
\end{figure}

In this paper, we propose a new technique called \name{} (Graph Recognition Convolutional Neural Network) that takes as input
flow charts as accurate specification,
 and uses deep convolutional neural network (CNN) to analyze a given image, and compiles the obtained information into program code.
\name{} is an end-to-end network that shares the computation of a rich convolutional feature vector and predicts edge and node information simultaneously. A flow chart is a diagram that represents the workflow of a program. It is used widely in  textbooks to teach coding and illustrate programs.  Moreover,  flow charts are  intuitive, which allows users to focus on programming logic instead of language details. Thus they are also frequently used in program designing stage.  
Figure \ref{fig:overview} shows a flow chart
representing the workflow  of a function that computes the absolute value of an input.
Because  flow charts can accurately describe programs,  synthesizing program by flow chart may precisely capture users' intent.

Due to recent progress in deep learning, CNN is powerful enough
to detect and recognize complicated information from image.
Therefore, we can obtain a graph data structure from a flow chart with the help of deep CNN.
Because neural networks are differentiable functions, our method does not suffer from the combinatorial explosion problem that plagues  traditional program synthesis methods.

Figure \ref{fig:overview} givens an overview of our approach.
First, an image of flow chart is resized to a fixed size and
fed to \name{}. Then, \name{}  generates the graph information including
an adjacent matrix for the edges and
a list of strings for the nodes.
Finally, we compile the graph information to source code.

Our evaluation showed that it is feasible to share the
convolutional vector between edge and node networks.
Experiments on our synthetic test dataset show that the accuracy rate to synthesize a  program is 66.4\%, and the accuracy rates to predict edges and nodes are 94.1\% and  67.9\%, respectively.
Experiments on another dataset, which is manually converted from a textbook, show that that the accuracy rates to synthesize a  program is 63.6\%, and the accuracy rates to predict edge and nodes are 72.7\% and  81.8\%, respectively.
The average time   to synthesize  a  program is about 60 milliseconds.

In summary, the main contributions include:
\begin{itemize}
\item{We propose a deep neural network that parses graph edge and node information from flow chart.
}
\item{We propose to use flow chart as an accurate and intuitive specification for program synthesis.}
\item{We have implemented a prototype and conducted empirical study.}
\end{itemize}

\begin{figure*}[t]
\centering
  \input{res/net.tikz}
  \caption{ Architecture of \name{}}
  \label{fig:net}
\end{figure*}

\section{Related Work}

Traditional methodology of program synthesis is to construct a
program space and design search algorithms to find a solution that satisfies the specification. The program space usually grows exponentially with the size of the target program. Different methods are proposed to speedup the search.
For example, Flashfill, \cite{gulwani2011automating},    synthesizes string-transforming programs given input-output examples. It uses dynamic programming to speedup the search.
Morpheus, \cite{wang2017synthesizing}, enumerates nested queries and prunes  by grouping programs with same input-output pairs.
Sketch, \cite{solar2008program}, and Sqlsol, \cite{cheng2019sqlsol}, encode the synthesis problem into logic constraints and
delegate the searching algorithm to modern SMT solvers.
This approach can  boost the performance because modern SMT solvers are implemented for efficiency.
Though different methods are proposed to speedup the synthesis process,
the underlying complexity is unchanged and scalability is still an issue
when it comes to large programs.
Another issue  is how to capture users' intent. 
Typically, synthesis algorithms  terminate when the first solution is found  or  the best-ranking solution is found. They may ask a user to provide more examples. However, there is no guarantee that the solution precisely captures the users' intent.

Recently researchers propose to use
machine learning to speed up the search for synthesized program.
Morpheus \cite{feng2017component} uses statistics to rank $R$ program sketches, and uses the rank to guide the search.
DeepCoder \cite{balog2016deepcoder}
augments beam search with deep learning recommendation, and the speedup is significant. However, the underlying complexity is still unchanged.

\ignore{Natural language based program synthesis takes as input a sentences in
natural language, and the synthesizer \emph{translates} it to a program, which conforms to the grammar of a DSL.}
Researchers proposed natural language based program synthesis techniques.
SQLizer, \cite{yaghmazadeh2017sqlizer}, synthesizes SQL queries from natural language.
Locascio et al, \cite{locascio2016neural}, synthesizes regular expressions from natural language.
\name{} differs from these approaches in that our input is different. Flow charts can accurately specify the users' intent, while 
natural language is ambiguous.

Faster RCNN, \cite{ren2015faster}, and LPRNet, \cite{zherzdev2018lprnet}, are closely related to our work.
Two subnets of \name{} are built following their ideas. Faster RCNN is a deep neural network for object detection.
It slides a small window on a convolutional feature and generates box proposals relative to anchors at each position.
The box proposal is used to crop the image for  a classifier to detect the class of the object in it.
Because of the shape and positioning of nodes in a flow chart are different from those in general object-detection jobs, \name{} chooses different anchors and  methods to select the proposals. LPRNet  is a license plate recognizing deep convolutional neural network.
It reads an image of a license plate and generates a sequence that preserves the spatial order of the characters.
A  feature of LPRNet is that it has only one deep convolutional network, while other work has both a CNN for feature extraction and a recurrent neuralnetwork for prediction.
This feature helps to limit the number of subnets in \name{}, since it already has four subnets.
Instead of taking a license plate as input, \name{} takes as input a crop of image whose boundary is predicted by another network.

\section{Network}

In this section, we  describe the  network and  its subnet.
Then, we describe the loss function and training details.

The input of \name{} is an image of a flow chart.
A flow chart is a graph diagram that represents the work flow of program~\cite{shelly2011discovering}.
Standard flow charts has several shapes for nodes.
In this work, we consider all shapes of nodes  as rectangles.
Because we use heuristic algorithms to generate ground truth data for  bounding boxes, designing and implementing algorithms for every shape requires considerable engineering effort . However, given ground truth data for bounding boxes, our algorithm can be trained the same way to handle other shapes.

The input flow chart is resized to a fixed size (400 pixels for height and 200 pixels for width)  before being fed to the neural network.
If both height and width is less then the fixed size, we pad
the image by zeroes. Otherwise, we interpolate the image to the fixed size. The output of \name{} is a graph representation of the flow chart, including an adjacent matrix for the edge representation and a list of text for content in nodes.
\ignore{
An adjacent matrix is a square matrix with values being 0, 1 or 2 and zeroes on its diagonal.
The element of the adjacent matrix at row $i$ and column $j$ has three possible values: 0, 1, 2, which encode no edge, normal edge or \texttt{YES} branch of a decision node, \texttt{NO} branch of a decision node, respectively.}
In addition to the original text, we insert an \emph{id} to the text
to match the text with the \emph{id}-th row of  adjacent matrix.
In this project, we use an alphabet of 50 characters including
English characters in lower case, digit characters, arithmetic operators
and other symbols.
The source code that our algorithm produces supports sequential statements
and control structure including IF-ELSE, WHILE loop, DO-WHILE loop.

Figure \ref{fig:net} is the overview of \name{}.
\name{} has four parts: backbone network, edge detection network,
node detection network, and node recognition network.
The backbone network takes  as input the raw image of the flow chart and outputs a feature vector.
The feature vector is fed to  both the edge detection to
produce the adjacent matrix of the flow chart,
and the node detection network to generate the bounding boxes of each node,
which is used to crop the node from the original image.
Then, the crop of each node  is fed to the node recognition
network to generate the text in the node.

\subsection{Backbone Network}
\label{sec:net_backbone}
The backbone network is a deep CNN that takes as input an image of size $C_{in} \times H \times W$. The output is a rich feature vector, which is later used as input to the edge network and the node detection network. The backbone network is a sequence of four basic blocks. 
Table \ref{tbl:basicblock} shows the architecture of a basic block, which takes as input a feature vector of $C_{in}$ channels and outputs a feature vector of $C_{out}$ channels and the same height and width.
Each convolutional layer in the basic block is followed by a batch
normalization layer and a ReLU activation layer. Each basic block is followed by a max-pooling layer, except the third one.
During training, a dropout layer (p=0.1) is added after each pooling layer.

Optionally, the backbone network can be made a residual learning network (ResNet), \cite{he2016deep}, by  modifying the basic block as follows.
We perform a down-sampling on the input vector with $1 \times 1$ convolutional layer to $C_{out}$ channels, and add the result to the original output as new output.

\ignore{
We use basic blocks of convolutional layers to construct the backbone
network.
A basic block network takes as input a tensor of shape $C_{in} \times H \times W$ and outputs a tensor of shape $C_{out} \times H \times W$.
Table \ref{tbl:basicblock} describes the architecture of a basic block
network.
Each convolutional layer in the basic block is followed by a batch
normalization layer and a ReLU activation layer.
In this paper, all Convolutional layers have padding values such that the output size does not change, unless explicitly mentioned.

Table \ref{tbl:backbone} describes the architecture of the backbone network
.
The backbone network can be made residual network by
adding the basic input to output.
During training, a dropout layer (p=0.1) is added after each pooling layer.

}

\begin{table}
  \caption{Basic Block Architecture.}
  \label{tbl:basicblock}
  \centering
  \begin{tabular}{|l|l|} \hline
    Layer Type & Parameters \\\hline
    Input  & $C_{in} \times H \times W$ \\\hline
    Conv2D & $C_{out} / 4, 3 \times 3$ \\\hline
    Conv2D & $C_{out} / 4, 3 \times 3$ \\\hline
    Conv2D & $C_{out} / 4, 3 \times 3$ \\\hline
    Conv2D & $C_{out} , 3 \times 3$ \\\hline
  \end{tabular}
\end{table}
    
\begin{table}
  \caption{Backbone Network Architecture}
  \label{tbl:backbone}
  \centering
 \begin{tabular}{|l|l|} \hline
    Layer Type & Parameters \\\hline
    Input & C $\times$ H $\times$ W \\\hline
    Basic Block & 15, 3 $\times$ 3 \\\hline
    Max Pool  & 2 $\times$ 2 \\\hline
    Basic Block & 50, 3 $\times$ 3 \\\hline
    Max Pool  & 2 $\times$ 2 \\\hline
    Basic Block & 200, 3 $\times$ 3 \\\hline
    Basic Block & 400, 3 $\times$ 3 \\\hline
    Max Pool  & 2 $\times$ 2 \\\hline
  \end{tabular}
\end{table}

\subsection{Edge Network}
\label{sec:net_edge}
The edge network takes as input the feature vector generated by the backbone network and outputs an adjacent matrix, which is the edge representation of the flow chart.
The element of the adjacent matrix at row $i$ and column $j$ has three possible values: 0, 1, 2, which encode no edge, normal edge or \texttt{YES} branch of a decision node, \texttt{NO} branch of a decision node from node $i$ to node $j$, respectively.
Because the number of nodes in the flow chart may vary, we pad the adjacent matrix to a fixed size $PAD$ by zeroes.
We set $PAD = 6$ in this work.
We encode the three values in the adjacent matrix with one-hot-vectors of length 3.
Therefore, the edge network outputs a vector of  $PAD \times PAD \times 3$ scores.

\ignore{
We model edge prediction as a multi-label classification problem.
The element at row $i$ and column $j$ in the adjacent matrix can have three values: 0, 1, 2.
The three values represents the there is no edge, normal edge or \texttt{YES} branch of a decision node, \texttt{NO} branch of a decision node, respectively.
We encode the tree values with one-hot vectors, and flatten the adjacent
into a vector of zeroes and ones, which is used as the label to train the
edge network. Note that \name{} accepts flow charts with various  the number of nodes, so the size of the adjacent matrix may vary.
We pad the adjacent matrix to a fixed size $PAD$ by zeroes.

The edge network takes as input the feature vector generated by the backbone network and outputs $PAD \times PAD \times 3$ scores.
}

Table \ref{tbl:edge} shows the architecture of the edge network.
It first performs a convolutional layer activated by a ReLU function.
Then, a max pooling layer is performed followed by two linear layers
actived by Tanh function.
We observed the training converges significantly faster when using Tanh activation function in the linear layers than using other activation functions.

The loss function to train the edge network is the multi-class
multi-classification hinge loss .
Equation \ref{equ:edge} is the formula of the loss function,  where x is the input vector and y is the target class indices.

\begin{equation}
  \label{equ:edge}
  l_{e}(x,y) = \sum_{i,j}\frac{max(0, 1-(x[y[j]]-x[i]))}{x.size(0)}
\end{equation}

\begin{table}
  \caption{Edge Network Architecture}
  \label{tbl:edge}
  \centering
 \begin{tabular}{|l|l|} \hline
    Layer Type & Parameters \\\hline
    Conv2D  & $400, 3 \times 3 $ \\\hline
    ReLu   &  \\\hline
    Max Pool  & $3 \times 3$ \\\hline
    Linear  & $400*16*7 \times 400 $ \\\hline
    Tanh  &   \\\hline
    Linear & $400 \times 6*6*3 $ \\\hline
    Tanh & \\\hline
  \end{tabular}
\end{table}

\subsection{Node Detection Network}
\label{sec:net_nd}
The node detection network takes as input the convolutional feature vector from the backbone network and outputs a set of rectangular node proposals, each with an objectness score.

The network slides a small network over the input convolutional layer. The output of the window is fed into two sibling layers: a box-regression layer and a box-classification layer. Because the small networks work in the sliding window fashion, they are naturally implemented with convolutional networks.

At each sliding window, we predict one region proposal which encodes the four coordinates of a box and one score which estimates the probability of whether the proposal is a node or not.
Each region proposal is parameterized relative to a reference box, called anchor. 
We parameterize the coordinates of the bounding boxes following \cite{girshick2014rich}:
\begin{equation*}
t_x = (x-x_a)/w_a,
t_y = (y-y_a)h_a,
\end{equation*}
\begin{equation*}
t_w=log(w/w_a),
t_h=log(h/h_a)
\end{equation*}
\begin{equation*}
t_x^*=(x^*-x_a)/w_a,
t_y^*=(y^*-y_a)/h_a,
\end{equation*}
\begin{equation*}
t_w^*=log(w^*/w_a),
t_h^*=log(h^*/h_a),
\end{equation*}
where $x$, $y$, $w$, and $h$ is the box’s center point and width and height. Variables $x$,  $x_a$ and $x^*$ are for the predicted box, anchor box, and ground-truth box respectively (likewise for $y$, $w$, $h$).
This parameterization converts large integers of bounding box coordinates to variables close to interval $[-1, 1]$, and therefore improves the numerical performance of the algorithm.

To train the node detection network, we assign a binary class label of being a node or not to each anchor.
We assign  a positive number to two kinds of anchors:
(1) The anchor with the highest Intersection-over-Union (IoU) with a ground-truth box, or
(2) an anchor that has an IoU higher than 0.9 with any ground-truth box.
We assign a negative number to an anchor if its IoU is less than 0.3 for all ground-truth boxes.
An anchor that is neither positive nor negative does not contribute to the train objective.
Because the number of positive labels and negative anchors  may be different and therefore the train may be biased toward one direction,  we sample from the more to ensure the same size of positive and negative labels during training.

We apply binary cross entropy loss to the objectness, and smooth-l1 loss
to the region proposal. The final loss is the sum of the objectness loss and region proposal loss over all anchors.

For prediction, we choose top 50 anchors with the highest scores,
and group them by the condition that anchors with IoU greater than 0.2 are in the same group. Then, we choose from each group the highest score as the final prediction.

Nevertheless, our network differs from the faster RCNN in two ways.
(1) Because the nodes in flow charts have similar size and shape, we use one anchor, instead many anchors with different size and ratio, to save computing power.
(2) Because the nodes in flow charts do not overlap, we  consider proposals in the same group if one has IoU over a threshold with any other one in the group.
Meanwhile, fast RCNN considers proposals in the same group if one has IoU over  a threshold with the one of highest score.

\ignore{

The architecture of the node detection network follows the idea of
faster RCNN \cite{}.

We define an anchor, which can map to a box in the original image, at each position of the feature map.
We use the anchor size of \lcnote{size}.

The network has a intermediate convulutional layer, a classifier, and
a regressor.
The intermediate convulutional layer takes as input the convulutional
feature vector,  and outputs a vector which is used by the classifier
and regressor separately.
The regressor is a one-channel convulutional layer which
generates one score at each position of the feature vector.
The score predicates the \emph{objectness} of the position, which is
whether this position maps to a node or background in the original image.
The classifier is a four-channel convulutional layer which
generates 4 coordinates at each position of the feature vector,
which regresses the bounding box if this position is a node.

We label the objectness using the IoU of the mapping box and the ground truth box of the node: If IoU greater than 0.9, we label it 1.
If IoU is less than 0.3, we label it 0. Points in between does not contribute to the loss.
Also, for each ground truth box, we label the highest IoU 1.
We random choose the same number of background points as the nodes,
and apply cross entropy loss to objectness classifier.

We use the transformation to the boxes, and apply the smooth L1 loss
to the regressor.

When prediction, we choose top 50, and then mms is applied.
The boxes with over 0.5 IoU are considered same.

}

The architecture of the node detection network is as follows.
The intermediate layer is a Conv2D layer with 3 X 3 kernel and  400 output channels.
The classifier  is a Conv2D layer with 3 X 3 kernel and 1 output channel.
The regressor  is a Conv2D layer with 3 X 3 kernel and  4 output channels.
Each Conv2D layer is activated by ReLU function.

\subsection{Node Recognition Network}
\label{sec:net_nr}
The node recognition network  takes as input
the crop of each node and outputs a vector of size $LENVOC \times 1 \times W$, where $LENVOC$ is the vocabulary size.

Table \ref{tbl:noderec} is the architecture of node recognition network,
where the base block shares the same architecture as the basic block in the backbone network, Table \ref{tbl:basicblock},

\ignore{
Following LPRNet, \cite{zherzdev2018lprnet},
The feature vector runs through a deep CNN, and
the output together with a global context
goes through another conv layer.
after pooling,  the input vector to height $1$ to get a sequence that preserves the spatial order of the text.}

\ignore{
Because the height of the output vector is pooled to 1 and the channel size is $LENVOC$,  at each position along the width axis, the channel values is interpreted as a probability distribution over the vocabulary.
 Since the output vector preserves the order of characters,
CTC loss \cite{graves2006connectionist} is applied to train the network.
}

The output of the node recognition network can be interpreted as
probability distribution over the vocabulary at each position along the width direction.
CTC loss is a  function for training sequences problems such as
handwriting recognition or speech recognition.
CTC loss does not attempt to learn the character boundaries, and
can be applied if the input is a sequence with some order.
By adoption CTC loss, we do not need to use another recurrent neural network to predict the text. Instead, we directly predict from the output of the node recognition network.

When prediction, we use greedy search to decode text from the output vector.

Optionally, a spatial transformer network (STN), \cite{jaderberg2015spatial}, can be inserted before the node recognition network to further adjust the the boundary of the crop.
 STN is the network that can be
inserted into existing convolutional architectures, giving neural networks the ability to actively spatially transform feature maps.

\ignore{
We follow the idea of LPRNet \cite{zherzdev2018lprnet} network,
and modified based on our dataset.
The node recognition network is another deep CNN.
Its architecture is a multi-layer of basic blocks interleaved by pooling
layers.

The key point here is that the height of the output vector is \emph{one}.
Therefore, it becomes a sequence which preserves the order of the characters. Therefore, a CTC loss can be applied. 
Therefore, No need to use an RNN.
}

\begin{table}
  \caption{Node Recognition Network Architecture}
  \label{tbl:noderec}
  \centering
 \begin{tabular}{|l|l|} \hline
    Layer Type & Parameters \\\hline
    Input & C $\times$ H $\times$ W \\\hline
    Basic Block & 64, 3 $\times$ 3 \\\hline
    Max Pool  & 3 $\times$ 3 \\\hline
    Basic Block & 128, 3 $\times$ 3 \\\hline
    Max Pool  & 3 $\times$ 3 \\\hline
    Basic Block & 256, 3 $\times$ 3 \\\hline
    Max Pool  & 3 $\times$ 3 \\\hline
    Basic Block & LENVOC, 3 $\times$ 3 \\\hline
  \end{tabular}
\end{table}

\subsection{Train and Implementation}

The whole network is  trained end-to-end. The lost function is the sum of loss of all sub-networks:
\begin{equation}
  loss = loss_{edge} + loss_{ndc} + loss_{ndr} + \sum_{node}{loss_{nr}},
\end{equation}
where $loss_{edge}$, $loss_{ndc}$,  $loss_{ndr}$ and $loss_{nr}$
is the loss of edge network,  classifier  and regressor of node detection network and node recognition network, respectively.

We use SGD method to train the network.
Learning rate is 0.02, and halved when the error plateaued.
The total epoch is 200.

\section{Experiment}

We designed experiments to answer the following research questions:
(1) what is the accuracy of \name{} ?
(2) what is the inference performance of \name{}?
(3) Is \name{} able to synthesize real-world program?

The experiments are conducted on a desktop with Geforce 1070 GPU,
Intel i7 CPU, and 16GB memory.
\name{} is implemented with PyTorch.

We tested the following networks.
\begin{itemize}
\item{\name{}: which is our basic network}
\item{Separated \name{}:  not sharing the backbone network. See Section \ref{sec:exp_sep}.}
\item{\name{} with ResNet: add optional residual learning. See Section \ref{sec:net_backbone}.}
\item{\name{} with STN: add option STN network. See Section \ref{sec:net_nr}.}
\item{\name{} with ResNet and STN.}
\end{itemize}

\subsection {Dataset Generation}
 We created a synthetic dataset to train and test the \name{}, since there are no existing datasets we can use.

A data sample includes a flow chart image in PNG format and a text file
 containing the ground truth which is used to train the network. The ground truth includes the adjacent matrix which represents the edge information,
and the bounding box and text content for each node.

The dataset contains flow charts with 3 to 6 nodes, and 0 to 2 decisions.
The text in each node contains 3 to 9 random characters from an alphabet of 50 characters, including the lower case English characters, digit characters, arithmetic operators, parenthesis et al.
Note that the alphabet can be chosen freely  without changing the essential network architecture.
The train and test dataset contains 9960 and 2490 data samples, respectively.

\ignore{
The process of generating a data sample is as follows.
First, We generate adjacent matrix of size from 3 to 6, with decision from 0 to 2 (inclusive);
and we generate a set of nodes, each node contains 3-9 random characters.
The characters are in the vocabulary of all lower English characters and some special tokens.
}

 We draw the flow chart using Graphviz, a popular graph-drawing tool.
We set the maximal width to be  200 pixels and maximal height to be 400 pixels.

When drawing the flow chart, we randomize to cover wide range of data samples.
The width of lines, including nodes boundaries and edges, is randomly chosen between 1 to 5 pixels.
The font size of characters is randomly chosen between 20 to 30.
The font color of characters is random RGB color.
We draw  nodes in rectangles and design a heuristical algorithm to effectively  compute high accuracy coordinates of bounding boxes. 

\ignore{Though we draw nodes in rectangle,
in standard flow charts, there are diamond and parallelogram.
We can train the same way given the bounding boxes.}

\subsection{Accuracy}

\begin{table}
  \caption{Accuracy in Percentage of \name{}. The columns are
  Edge, Sequence, Nodes and Graph accuracy, respectively.}
  \label{tbl:acc}
        \centering
  \begin{tabular}{lcccc}\toprule
               & Edge & Squence & Nodes & Graph \\             \midrule
   \name{}       & 94.1 & 90.6    & 67.9  & 66.4 \\\midrule
   +STN        & 90.0 & 93.8    & 71.3  & 64.4  \\
   +ResNet     & 91.9 & 91.3    & 64.8  & 60.3  \\
   +STN+ResNet & 90.9 & 93.2    & 70.0  & 63.2  \\ \midrule
   Separated   & 93.7 & 91.1    & 68.2  & 65.8 \\   \bottomrule
\end{tabular}
\end{table}

\begin{table}
  \caption{Time Cost in Milliseconds of \name{} . The columns are the time cost of the
  backbone, edge, node detection, node recognition network and \name{}, respectively.}
  \label{tbl:time}
\centering
  \begin{tabular}{lccccc}\toprule
        & BB & Edge & ND  & NR & \name{} \\\midrule
   \name{} & 3.5      & 0.4  & 42.4 & 14.2 & 60.5 \\ \midrule
   +STN       & 3.5      & 0.4  & 43.3 & 16.6 & 63.8 \\
   +ResNet    & 3.5      & 0.4  & 44.8 & 14.0 & 79.1 \\
   +STN+ResNet& 3.6      & 0.5  & 47.0 & 16.7 & 84.3 \\   \midrule
   Separated  & 3.5      & 0.4  & 42.8 & 15.1 & 61.8   \\\bottomrule
  \end{tabular}
\end{table}

\begin{figure*}[t!]
\centering
\begin{tabular}{|cl|cl|} \hline
  \begin{minipage}{0.2\textwidth}
    \includegraphics[scale=0.5]{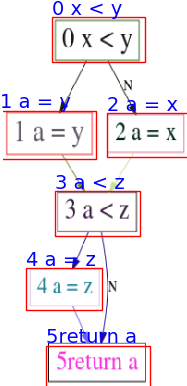}
    \end{minipage}
  &
  \begin{minipage}{0.25\textwidth}
    {\scriptsize
\begin{verbatim}
[[0 1 2 0 0 0]
 [0 0 0 1 0 0]
 [0 0 0 1 0 0]
 [0 0 0 0 1 2]
 [0 0 0 0 0 1]
 [0 0 0 0 0 0]]
\end{verbatim}
    \vspace{1em}
\begin{verbatim}
if x<y
    a=y
else
    a=x
if a<z
    a=z
return a
\end{verbatim}
    }
  \end{minipage}
  &  
    \begin{minipage}{0.2\textwidth}
    \includegraphics[scale=0.5]{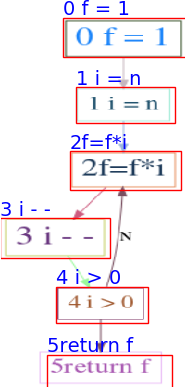}
    \end{minipage}
  &
  \begin{minipage}{0.25\textwidth}
    {\scriptsize
\begin{verbatim}
[[0 1 0 0 0 0]
 [0 0 1 0 0 0]
 [0 0 0 1 0 0]
 [0 0 0 0 1 0]
 [0 0 2 0 0 1]
 [0 0 0 0 0 0]]
\end{verbatim}
    \vspace{1em}
\begin{verbatim}
f=1
i=n
do
    f=f*i
    i--
while i>0
return f
\end{verbatim}
    }
    \end{minipage}

  \\\hline
\end{tabular} 
\caption{Samples of \name{} prediction. The  predicted bounding boxes are drawn in red lines. The predicted text is drawn above the bounding box in blue characters.  On the right  is the predicted adjacent matrix and synthesized source code.}
\label{fig:sample}
\end{figure*}

We measure the Edge Accuracy, Sequence Accuracy, Nodes accuracy and Graph Accuracy  of \name{} prediction.
Edge accuracy is the percentage of images whose edge is correctly predicted.
We say the edge is correctly predicted if the predicted adjacent matrix is exactly the same as the ground truth.
Sequence accuracy is the percentage of nodes whose text content is correctly predicted among all nodes in all images.
This item shows how well the node recognition network works for individual nodes.
Nodes accuracy is the percentage of images whose nodes are all correctly predicted.
This item shows how well the tool predicts the nodes as a whole.
Graph accuracy is the percentage of  images whose edge and nodes are all correctly predicted.

The accuracy of \name{} and its subnets are described in the first row of Table \ref{tbl:acc}.
The Edge, Sequence, Nodes and Graph accuracy is 94.1\%, 90.6\%, 67.9\% and 66.4\%, respectively.
Note that the Graph accuracy is the result of joint probability of
all edges and all nodes, so it is lower than Edge and Sequence accuracy.

We conducted ablation study to identify how well the optional enhancements are.
In Table \ref{tbl:acc}, row 2, 3 and 4 shows the accuracy of \name{} with different enhancements.
We find that the edge accuracy and graph accuracy decreased with those enhancements. Our explanation is that the extra trainable weights in the enhancements caused the network biased toward the nodes detection and recognition networks, but caused the decrease in edge accuracy, and therefore caused decrease in the graph accuracy.
 We also find all STN enhancements improved the accuracy of node detection and recognition networks.

\subsection{Performance}
Table \ref{tbl:time} shows the performance of \name{} and its subnets.
The overall time cost of \name{} is about 60 milliseconds, 
and the performance of other networks are close to the performance of \name{}.
Among the subnets, the time cost of node detection network is the major part, 70.1\%,  of the overall cost.
The rest is the node recognition network (23.5\%), backbone network (5.8\%), and edge network (1\%).

\subsection{End-to-End vs Separated Network}
\label{sec:exp_sep}
Because edges and nodes information
do not depend on each other,
it is natural to consider to use two separate networks to predict
edges and nodes.
We designed experiments to see how well both ideas work.

We made two clones of \name{} and modified as follows.
For one clone, we disable the node detection network and node recognition
network.
For the other clone, we disable the edge detection network.
We trained the two networks separately with the same dataset and hyper parameters.

The row \emph{Separated} in Table \ref{tbl:acc} and \ref{tbl:time}
shows the accuracy and performance of the separated networks.
The performance are close to \name{}, therefore it is feasible to share
 the computation of the backbone network.
The sharing  saves the 3.5 milliseconds, which is 5.8\% of the time cost.

\subsection{Real-world Program Synthesis}

\begin{table}
  \caption{Result of real-world program synthesis. The columns are program name, Graph accuracy, Edge accuracy, Nodes accuracy, number of nodes, correctly predicted nodes. }
  \label{tbl:sample}
  \centering
\begin{tabular}{lccccc}
\toprule
  program    & Graph & Edge & Nodes & \#Nodes & Nodes\\ \midrule
  abs        & 1   & 1      & 1        & 4         & 4   \\
  swap       & 1   & 1      & 1        & 3         & 3   \\
  max        & 0   & 1      & 0        & 4         & 3   \\
  sum        & 0   & 0      & 1        & 6         & 6   \\
  max3       & 1   & 1      & 1        & 6         & 6   \\
  log        & 1   & 1      & 1        & 4         & 4   \\
  radius     & 1   & 1      & 1        & 3         & 3   \\
  poly       & 1   & 1      & 1        & 6         & 6   \\
  factorial  & 1   & 1      & 1        & 6         & 6   \\
  quadrant   & 0   & 0      & 1        & 5         & 5   \\
  cntpos     & 0   & 0      & 0        & 6         & 3   \\\bottomrule
  sum        & 7   & 8      & 9        & 53        & 49  \\
  percentage & 63.6& 72.7   & 81.8     & -         & 92.5\\
\bottomrule
\end{tabular}
\end{table}

In order to see how well \name{} synthesizes real-world programs,
we created a test dataset of 11 programs, which are selected from the problems in a programming textbook \cite{cbook}.

Table \ref{tbl:sample} is the testing result.
It shows that 70\% of programs are correctly predicted,
70\% edges are correctly predicted,
93.6\% individual nodes are  correctly predicted.

In Figure \ref{fig:sample}, we draw  two samples to visually 
demonstrate  the input and output of \name{}.
The first one is the function that finds the max of three numbers.
The second one is the \texttt{factorial} function.

\section{Conclusion}
We presented \name{}, a deep convolutional neural network that
parses graph data structure from a flow chart,
and we automatically generates source code that matches the flow chart.
\name{} predicts the edge information 
and nodes information simultaneously.
Experiments show that we can share the computation of the feature vector.
\name{} achieves 66.4\%, 94.1\%, 90.6\% for graph, edge and node accuracy, respectively, on our test dataset, and close accuracy on a real-world dataset.

\bibliographystyle{IEEEtran}
\bibliography{flowchart}

\end{document}